# No country for old linguists: LLM–brain alignment underdetermines neural computation

Elliot Murphy[1]

1. Vivian L. Smith Department of Neurosurgery, McGovern Medical School at UTHealth Houston, Houston, Texas, USA

**Correspondence**: Elliot Murphy, Vivian L. Smith Department of Neurosurgery, McGovern Medical School, UTHealth Houston, Houston, Texas, USA

Email: elliot.murphy@uth.tmc.edu

## 1 | INTRODUCTION

Nastase et al. (2026) argue that large language models (LLMs) may illuminate language processing because both rely on distributed, context-sensitive representations shaped by statistical learning. Their rejection of simple cortical "boxology" is persuasive, and they articulate a strong case for the value of LLM–brain alignment research. The key question is what kind of inference LLM–brain alignment licenses.

My claim here will be narrow: representational alignment can in principle constrain mechanistic hypotheses, but it does not by itself identify a mechanism. Nastase et al. acknowledge that an encoding model can capture features represented in neural activity without establishing a shared architecture or algorithm. Yet the authors sometime move from alignment to "shared computational principles" and ultimately to LLMs as mechanistic models of natural language. Indeed, their methodological caveat that alignment does not establish a shared architecture or algorithm sits uneasily with their conclusion that LLMs might instantiate the same computational principles as biological brains and provide a "fully mechanistic model" of language. Below, I discuss what I consider to be problems of *logical, causal, and computational underdetermination* in Nastase et al.'s (2026) proposal.

## 2 | FROM ALIGNMENT TO MECHANISM

A standard neural encoding analysis can be written schematically as

$$Y_{\text{brain}} = WX_{\text{LLM}} + \varepsilon,$$

where $X_{\text{LLM}}$ is an LLM representation, $Y_{\text{brain}}$ a neural response, W a fitted transformation, and ε residual variance. Held-out prediction shows that information in $X_{\text{LLM}}$ predicts $Y_{\text{brain}}$. It does not, on its own, establish that the two representational spaces share a geometry, since an unrestricted W can distort the relations at issue; in general,

$$\|Wx_i - Wx_j\|_2 \neq \|x_i - x_j\|_2,$$

unless W is constrained to preserve the relevant relations. Nor does generalization rescue the inference. The strongest geometric evidence in this recent literature – zero-shot prediction of neural activity for left-out words under a linear map fitted on others – shows that the fitted mapping generalizes across the lexicon, and therefore that a predictive linear relation exists between the two spaces over held-out items. But an unconstrained W is free to shear, rescale, and collapse dimensions: the relations guaranteed to survive it are those invariant under arbitrary linear maps, such as collinearity and parallelism, not the distances, angles, and neighborhood structure that "geometry" is ordinarily taken to mean. Mappings constrained to preserve distances, or comparisons of similarity structure computed within each system separately, escape this particular objection – but they still inherit the problems that follow. Even granting shared geometry in the strongest sense, establishing that two systems occupy similar representational geometries fixes neither the operations that produced those geometries nor the operations defined over them.

This point is sharpened by a common-input problem. Both $X_{LLM}$ and $Y_{brain}$ are responses to the same structured linguistic stimulus *S*. Thus,

$$\mathrm{S} \rightarrow \mathrm{X}_{\mathrm{LLM}}$$
$$\mathrm{S} \rightarrow \mathrm{Y}_{\mathrm{brain}}$$

can generate alignment even when the systems share no internal operation. Apparent alignment may therefore be inherited from shared stimulus structure rather than shared computation, a particular concern in naturalistic encoding designs (Schönmann et al., 2026). Guest and Martin (2023) formalize this problem clearly: Let Q be the claim that a model captures the relevant human mechanism, and P the observation that it predicts human behavioral or neural data. Q may imply P, but observing P and concluding Q gives:

$$Q \rightarrow P, \qquad P, \qquad \therefore Q.$$

This is – infamously – affirming the consequent. Nastase et al. (2026) recognize this objection and retreat to what they call a "narrower" and "more defensible" claim: "Both systems converge on a similar representational format: a distributed, high-dimensional population code for solving the same underlying learning problem". But this narrower claim is not secured by the evidence either. Convergence on a representational format is a claim about shared geometry, and the regression-based results mentioned above establish only that a predictive linear relation exists between the two spaces – a relation equally compatible with formats whose distances, angles, and neighborhood structure disagree. The retreat from mechanism to format does not shorten the inferential distance; it merely relocates it. Prediction, alignment, shared geometry, shared computational principles, and mechanism remain distinct claims, and a regression coefficient alone cannot supply the transitions between them.

## 3 | LOST IN TRANSLATION: A LESSON FROM SURPRISAL

Surprisal provides a useful analogy here. Lexical surprisal is:

$$I(w_i \mid c_i) = -\log p(w_i \mid c_i),$$

and often predicts processing difficulty and neural responses extremely well. Yet, Slaats and Martin (2025) show that syntactic structure can alter surprisal while surprisal itself fails to identify that latent structure; part of its variance can also reflect simpler distributional properties. A strong descriptor need not be a mechanism.

The same caution applies to contextual embeddings. Their strength is that they absorb variance associated with lexical identity, frequency, syntax, semantics, discourse, pragmatics, and world knowledge. That breadth makes them powerful predictors but can obscure the source of neural predictivity. Predictive coverage is not causal discrimination, and a small brave few have questioned the claim that LLMs have mastered human language (Murphy et al., 2026). No doubt distributional information can be an important *cue*, but the point here is that a cue is not the operation it constrains:

an embedding can reveal stimulus dimensions represented in the brain without identifying the biological transformation that implements composition.

## 4 | THE MAP IS NOT THE MECHANISM

These observations matter because Nastase et al. repeatedly invoke neural mechanisms, yet the clearest mechanisms in their examples are inside artificial models. Attention heads, weight matrices, and nonlinearities causally transform one LLM state into another. Consider that Kumar et al. (2024) show that BERT attention-head transformations predict fMRI activity across the language network, but the decomposed "circuitry" is BERT circuitry: the analysis does not identify the corresponding cortical operation. Likewise, Zada et al. (2024) show that GPT-2 embeddings capture linguistic information shared between speakers and listeners, not the neuronal transformation that produces or transmits it.

A stronger mechanistic claim would specify a candidate operation, identify a biological organization capable of implementing it, derive a discriminating prediction, and ideally perturb the relevant process. Neural measurements are evidence for mechanistic accounts, not mechanisms themselves (Ross & Bassett, 2024; van Bree, 2024).

## 5 | WITH GREAT DIMENSIONALITY COMES GREAT RESPONSIBILITY: POPULATION CODES NEED COMPUTATIONAL CONSTRAINTS

Population codes and mixed selectivity are widespread, making "high-dimensional population code" a representational format rather than a theory of a particular linguistic computation. LLM–brain alignment has therefore narrowed the space of neural *descriptions* more than the space of neural *mechanisms*. Population geometry, manifolds and representational subspaces provide geometric or topological descriptions; absent a linking account, such descriptions remain non-mechanistic and may be causally agnostic mathematical models rather than accounts of the processes generating the observed states (Figure 1; see Ross & Bassett, 2024; van Bree, 2024). Thus, showing where linguistic information lies in neural state space does not explain which causal neural operations generate, transform or retrieve it. Overlapping populations can multiplex variables across subspaces, time windows, trajectories, oscillatory regimes, or coupling patterns. In contrast, formal properties of language can narrow candidate explanatory structures by ruling out implementations that fail to preserve the required linguistic relations (Figure 1).

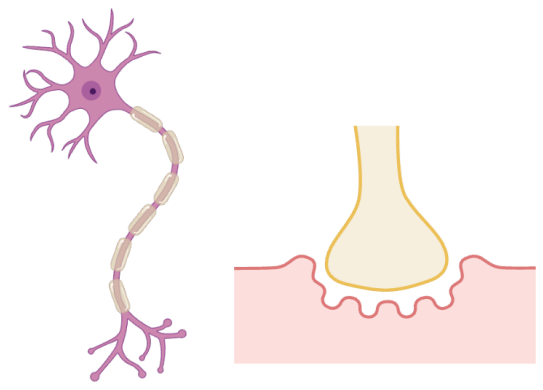
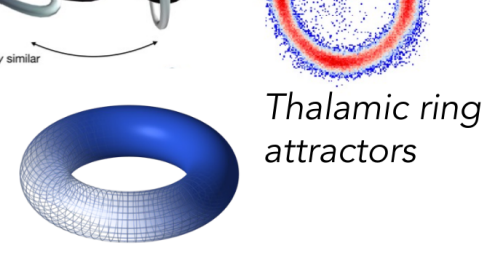
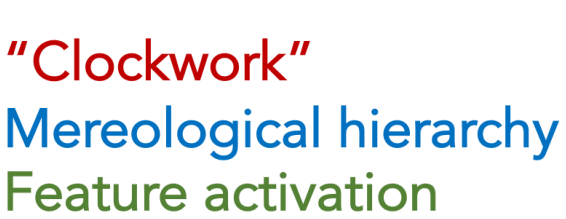

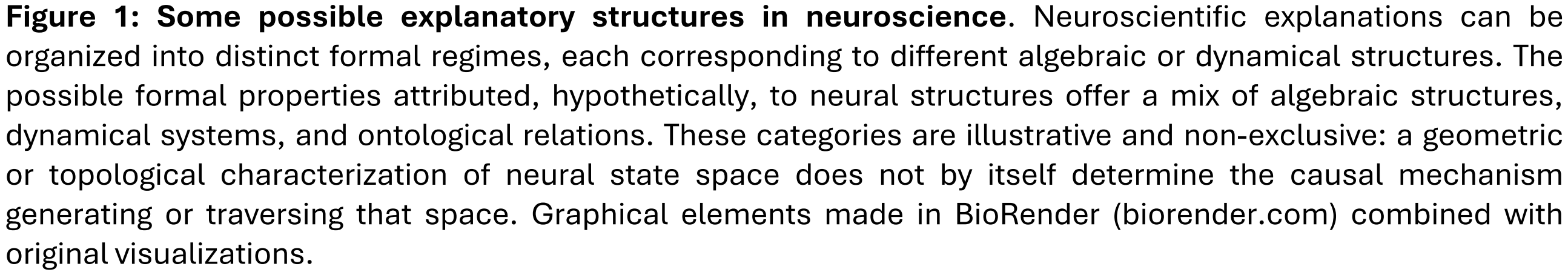

**Figure 1: Some possible explanatory structures in neuroscience**. Neuroscientific explanations can be organized into distinct formal regimes, each corresponding to different algebraic or dynamical structures. The possible formal properties attributed, hypothetically, to neural structures offer a mix of algebraic structures, dynamical systems, and ontological relations. These categories are illustrative and non-exclusive: a geometric or topological characterization of neural state space does not by itself determine the causal mechanism generating or traversing that space. Graphical elements made in BioRender (biorender.com) combined with original visualizations.

A further distinction is between information being decodable from a population and being computationally addressable by the system itself. Long-distance dependencies, agreement, ellipsis and structural reanalysis require already-built constituents to be selectively retrieved and reused. A high-dimensional sentence representation may preserve enough information for an external decoder to recover such structure while providing no internal operation that can reopen one constituent without reinstating the whole state. Decodability therefore establishes informational availability to the analyst, not functional accessibility to the brain's linguistic computation. A population code may thus be representationally rich yet computationally insufficient for the operation being explained.

Nastase et al. list several such differences among the "conspicuous gaps" between artificial and biological networks: transformers retain veridical access to thousands of tokens of prior context, whereas humans must immediately condense the present input to be ready for the next, and text-based models lack the endogenous oscillatory dynamics through which cortex is thought to schedule, maintain and release linguistic structure. They present these as engineering shortfalls, to be closed by better architectures, but a system with veridical access to its entire context never confronts the same bottleneck, so its success cannot by itself reveal the brain's solution to it. Such constraints are not nuisance parameters to be abstracted away; they are part of what any admissible implementation must satisfy (Murphy, 2025).

There is also a falsifiability problem that emerges here. A sufficiently high-dimensional population code with mixed selectivity can accommodate overlap, dissociation, multiplexing, and context dependence. If almost any neural pattern can be redescribed as activity in some subspace, what result would count against the proposal? Without forbidden outcomes, "population code" risks naming an implementation vocabulary rather than a computational theory (Guest, 2024). Nor is vectorial computation necessarily opposed to algebraic theory (Marcolli et al., 2025; Murphy, 2025). Algebra need not posit literal symbols or trees in cortex; it can specify invariants that any implementation, including continuous high-dimensional dynamics, must preserve.

The issue, then, is not whether linguistic structure *can* be embedded in a vector space, but whether the operations over that space preserve the relations that define the structure. Ordinary geometric similarity, vector addition, or averaging provide no such guarantee. For syntax, for example, composition is non-associative: ((a ⋆ b) ⋆ c) need not equal (a ⋆ (b ⋆ c)). A representation that collapses these alternatives into the same or indistinguishable population state has erased precisely the constituent structure that the computation must preserve. Similar problems arise for asymmetric entailment, scope, binding, and type-sensitive semantic composition. Thus, showing that linguistic information occupies recoverable directions or subspaces does not establish that the neural dynamics possess the algebraic operations required to *use* that information compositionally. While Nastase et al. (2026) claim that "the formal constructs we have developed to describe linguistic structure stubbornly refuse to map onto the neural code", they do not engage with neurocomputational models that are able to show partial success here (e.g., Murphy, 2025).

An approach centered on neural admissibility of formal properties can emerge here. Let M be a formally specified cognitive operation, N a mapping from cognitive objects to neural states, and G a candidate neural transformation. An admissible implementation should satisfy a structure-preservation (homomorphism) condition:

$$\mathcal{N}\big(\mathcal{M}(x,y)\big) \;\simeq\; G\big(\mathcal{N}(x),\mathcal{N}(y)\big),$$

where the relevant equivalence relation must be independently justified. The aim here is not exact identity between mathematical and biological objects, but preservation of the relations defining the target capacity. For language, these may include hierarchical grouping, non-associative bracketing, recursive closure, structured access to subparts, headedness, or type compatibility (Murphy, 2025). This changes the focal research question from "can a decoder recover X?" to "can the proposed dynamics perform the transformation X requires?" A model may explain substantial neural variance yet fail that test. Formal theory can therefore eliminate candidate implementations rather than merely add another correlated label – providing a clear program to address Poeppel's mapping problem, rather than rhetorically dissolving it into population geometry.

## 6 | CONCLUSION

Nastase et al.'s (2026) contribution is important and timely. Distributed context-sensitive coding is more promising than one-to-one mappings between linguistic labels and cortical regions. The disagreement I have raised here concerns evidential level: predictive and representational success should constrain explanation, not count as explanation itself. Nastase et al. themselves propose the right criterion: a computational theory should illuminate neural mechanisms and generate discriminating, testable predictions. The LLM–brain literature has produced important clues and powerful descriptions of representational organization. Yet, the advantage of formally explicit symbolic–algebraic models is precisely that they specify which relations a candidate neural mechanism must preserve, thereby ruling out whole classes of implementations: prediction and correlation can describe neural activity, but explaining a causal mechanism requires showing how the relevant linguistic computation undergoes neural enforcement.

**CONFLICT OF INTEREST STATEMENT**

The author declares no competing interests.

**DATA AVAILABILITY STATEMENT**

Data sharing is not applicable to this article as no datasets were generated or analyzed.